\documentclass[conference]{IEEEtran}
\IEEEoverridecommandlockouts
\usepackage{cite}
\usepackage{amsmath,amssymb,amsfonts}
\usepackage{algorithmic}
\usepackage{graphicx}
\usepackage{textcomp}
\usepackage{xcolor}
\usepackage{fancyhdr}
\usepackage{colortbl}
\usepackage[colorlinks,urlcolor=blue,linkcolor=blue,citecolor=blue]{hyperref}
\usepackage{lipsum}% generate text for the example

\def\BibTeX{{\rm B\kern-.05em{\sc i\kern-.025em b}\kern-.08em
    T\kern-.1667em\lower.7ex\hbox{E}\kern-.125emX}}
    
\usepackage{tgpagella}
\usepackage{balance}

\begin{document}
\title{Toward Carbon-Neutral Human AI: Rethinking Data, Computation, and Learning Paradigms for Sustainable Intelligence%Human AI: Learning Every Day for a Sustainable and Carbon-Neutral AI Solutions
}

\author{
\IEEEauthorblockN{KC Santosh, Rodrigue Rizk, Longwei Wang}
\IEEEauthorblockA{
AI Research Lab, Department of Computer Science, University of South Dakota\\
414 E Clark St, Vermillion, SD 57069\\
Web: \url{https://ai-research-lab.org}, Email: {\{kc.santosh, rodrigue.rizk, longwei.wang\}@usd.edu}
}
}
\maketitle
\begin{abstract}
%\todo{check the commented abstract if u prefer it over the old one}
The rapid advancement of Artificial Intelligence (AI) has led to unprecedented computational demands, raising significant environmental and ethical concerns. This paper critiques the prevailing reliance on large-scale, static datasets and monolithic training paradigms, advocating for a shift toward human-inspired, sustainable AI solutions. We introduce a novel framework, Human AI (HAI), which emphasizes incremental learning, carbon-aware optimization, and human-in-the-loop collaboration to enhance adaptability, efficiency, and accountability. By drawing parallels with biological cognition and leveraging dynamic architectures, HAI seeks to balance performance with ecological responsibility. We detail the theoretical foundations, system design, and operational principles that enable AI to learn continuously and contextually while minimizing carbon footprints and human annotation costs. Our approach addresses pressing challenges in active learning, continual adaptation, and energy-efficient model deployment, offering a pathway toward responsible, human-centered artificial intelligence\footnote{TEDx Talk: \href{https://www.youtube.com/watch?v=J9dZV2EAuUU}{https://www.youtube.com/watch?v=J9dZV2EAuUU} --- Sustainable AI solutions}.
\end{abstract}

\begin{IEEEkeywords}
Green Computing, No-to-Carbon Footprint, Human AI, Sustainable AI
\end{IEEEkeywords}

\section{Introduction}

Artificial Intelligence (AI) has undergone unprecedented growth in the past decade, with state-of-the-art models achieving remarkable breakthroughs across domains such as natural language processing, computer vision, drug discovery, and climate modeling. However, this rapid progress comes at a substantial environmental cost. Training a single large language model can emit as much carbon as five cars in their lifetimes, with energy demands doubling approximately every few months in pursuit of marginal accuracy improvements ~\cite{Strubell2019EnergyAP, Patterson2021CarbonEA}.

While the current AI paradigm largely emphasizes scale, \textit{i.e.}, more data, bigger models, and higher compute budgets, emerging research suggests that more sustainable solutions/paths are not only possible but necessary. In particular, the reliance on large, indiscriminately collected datasets is increasingly being challenged. Instead, methods that prioritize data quality over quantity, such as meta-learning \cite{Vettoruzzo2023AdvancesAC}, active learning (AL) \cite{Santosh2023ActiveLT, Nakarmi2023ActiveLT}, and human-in-the-loop (HITL) systems \cite{Amershi2014PowerTT} have demonstrated the capacity to achieve comparable or superior performance under resource constraints. Moreover, the COVID-19 pandemic, for example, underscored the need for \textit{agile learning systems} capable of adapting rapidly to limited, evolving data. During this period, traditional data pipelines proved too slow, and high-performance models trained on outdated data became liabilities rather than assets. Human-expert-guided machine learning systems proved far more effective, revealing the potential for a new paradigm: one where incremental, context-aware, and energy-efficient learning is the default.

This paper proposes a foundational shift in the way we conceive of AI systems, from monolithic, carbon-intensive models to Human AI (HAI) systems that mirror human cognition: learning continuously, selectively, and responsibly. Such systems incorporate minimal computational waste, contextual adaptability, and deep integration of human knowledge.

We argue that sustainable AI is not merely an environmental imperative but a computational and epistemological one. By rethinking learning paradigms and embracing Green AI \cite{Schwartz2020GreenA}, we can ensure that future AI systems are not only powerful but ethical, explainable, and aligned with long-term societal values.

% This paper makes the following key contributions:
% We present a comprehensive survey of low-footprint AI paradigms and identify their common principles.
% We introduce a novel architectural blueprint for Human AI systems that combine meta-learning, continual learning, and human-in-the-loop supervision.
% We propose governance-aligned metrics for responsible AI deployment, rooted in explainability and sustainability.

\section{Green AI and the Carbon Footprint Associated With Machine Learning Models}
Estimating the carbon footprint of Machine Learning (ML) models involves quantifying energy consumption across training and inference phases and translating this into equivalent carbon dioxide emissions (\(\text{CO}_2\)e). A widely adopted formulation expresses the total carbon footprint \(C\) (in kg~\(\text{CO}_2\)e) as:
\begin{align}
&C = E \times CI = (P \times T) \times CI, 
\end{align}
where \(E\) denotes total energy consumption (in kWh), \(P\) is the average power draw of the system (in kW), \(T\) is the total runtime (in hours), and \(CI\) represents the carbon intensity of electricity (in kg~\(\text{CO}_2\)/kWh). This model underscores how both hardware efficiency and geographic location, specifically the energy mix of the local grid, influence overall emissions. The increasing computational demands of modern ML models have raised critical concerns regarding their environmental sustainability. For instance, \cite{Strubell2019EnergyAP} estimated that training a single transformer-based NLP model could emit over 626,000 pounds of CO$_2$, largely due to the energy-intensive GPU clusters required. Similarly, \cite{Patterson2021CarbonEA} quantified the emissions of Google’s deep learning workloads and emphasized the need for carbon-aware scheduling and hardware optimization.

The carbon footprint of ML workflows is governed by a range of interdependent factors: (a) model size and architectural complexity, (b) hardware efficiency and utilization, (c) total training and inference time, (d) energy source of the deployment region, (e) data center cooling overhead, and (f) software-level optimizations (e.g., compiler efficiency, mixed-precision training). Large-scale models, particularly those with billions of parameters, incur substantial computational demands, which are further amplified by the need to process and store vast datasets. This results in disproportionately high emissions, particularly when deployed in regions reliant on fossil fuel-based energy.

To address this, \cite{Schwartz2020GreenA} introduced the notion of \textit{Green AI}, advocating for efficiency as a first-order metric alongside accuracy. They argue for reporting energy usage and carbon emissions as standard practice in ML tools. Other efforts, such as CodeCarbon and ML CO$_2$ Impact calculators \cite{Lacoste2019QuantifyingTC}, have enabled more transparent reporting of energy consumption. However, such tools are often used retroactively rather than proactively guiding design.

Despite these efforts, the dominant paradigm still rewards scale, \textit{e.g.}, GPT-4 reportedly trained with over $10^6$ GPU hours. This paper challenges this trajectory by promoting \textit{low-carbon alternatives} centered on \textit{data efficiency}, \textit{incremental learning}, and \textit{human-AI symbiosis}.

%Recent tools such as \href{https://codecarbon.io}{CodeCarbon} and \href{https://github.com/romainzk/eco2ai}{eco2AI} provide practical frameworks to monitor and estimate ML energy usage. These tools incorporate hardware specifications, geolocation-based carbon intensity, and real-time energy metrics to offer accurate and actionable insights into the environmental cost of model development. By enabling transparent reporting and encouraging low-carbon design choices, they serve as foundational components in the shift toward sustainable AI.

%To address these challenges, the emerging field of sustainable machine learning advocates for data- and energy-efficient alternatives. Approaches such as active learning, continual learning, and meta-learning reduce reliance on large static datasets and instead favor incremental, task-adaptive updates. These methods not only lower computational overhead but also align with carbon-aware AI practices by minimizing unnecessary training cycles and focusing learning efforts on the most informative examples. In this context, carbon footprint minimization is not merely an engineering constraint but a central design objective that must be explicitly modeled and optimized in the next generation of AI systems.

\section{The Myth of Big Data and the Cost of Waiting}
A prevailing assumption in AI is that larger datasets invariably lead to better models. While scale can enhance generalization in some domains, recent advances in meta-learning and adversarial robustness challenge this belief, demonstrating that smaller, high-quality datasets can be more informative and efficient for guiding the learning process \cite{Vettoruzzo2023AdvancesAC, Jain2024NonuniformIA}. Moreover, despite the ubiquity of the term `big data,' the field lacks a principled definition of how much data is `enough' to begin solving real-world problems (\textit{`How big is Big Data?')}\cite{Santosh2021Covid19IT}.

This myth was starkly exposed during the COVID-19 pandemic. Delays in collecting large-scale, labeled datasets significantly hampered the responsiveness of machine learning systems. In contrast, approaches grounded in active learning and HITL paradigms enabled early detection and decision-making from limited but evolving data streams \cite{Santosh2023ActiveLT, Singh2025PATLPA, Nakarmi2023ActiveLT, Bouguelia2017AgreeingTD}. These methods leveraged context, uncertainty, and expert feedback to compensate for data scarcity, highlighting the critical value of adaptability over brute-force data accumulation. This raises a fundamental question: \textit{Are we truly solving problems if we wait years to gather `enough' data?} If AI systems are designed to operate solely on static, pre-collected datasets, they fail to reflect the dynamic nature of the real world. Effective AI must be built not just for accuracy, but for immediacy, capable of learning in real time, adapting to changing conditions, and responding to high-stakes scenarios as they unfold.

The next global epidemic is not a matter of \textit{if}, but \textit{when}. If AI is to function as an early warning system, rather than merely a post-crisis analyst; it must be designed to learn as humans do: incrementally, contextually, and continuously. The goal is not to collect the largest dataset, but to learn from all available, relevant cases as they emerge.

\section{Learning Every Day: A Human Model for Machines}
%\subsection{Cognitive and Continual Learning Paradigms}

%The convergence of these research streams, \textit{i.e.,} Green AI, meta-learning, HITL frameworks, and continual learning, points toward a new, underexplored paradigm: incremental, sustainable, human-centered AI. This work synthesizes these threads into a unified framework, proposing a shift from model-centric to system-centric AI design that internalizes \textit{not just what AI learns, but how and at what cost}.

Unlike conventional ML models that rely on episodic, batch-based training, human learning is inherently continuous and incremental, beginning from birth. Individuals do not wait to accumulate large volumes of data before updating their knowledge; instead, they assimilate new information daily, in cognitively manageable portions. Unlike current AI systems, which often require full retraining to adapt to new tasks, human cognition exhibits \textit{lifelong learning}, continually integrating new information without catastrophic forgetting. Standard AI workflows typically involve waiting for sufficient data accumulation before retraining models from scratch, a process that is both time-consuming and energy-intensive, contributing significantly to the carbon footprint of large-scale AI. A more sustainable alternative is to design systems that “learn small things every day,” reducing both computational cost and environmental impact. Emulating this capacity is the goal of continual learning (CL), which studies how models can retain and accumulate knowledge over time. This context-sensitive and ongoing engagement supports real-time adaptation, reduces cognitive load, and yields long-term efficiency. By aligning AI systems with this natural paradigm, future models can become more responsive, adaptive, and sustainable, while better reflecting the underlying mechanisms of human cognition. 

Catastrophic forgetting, first observed in neural networks by \cite{McCloskey1989CatastrophicII}, has been addressed through methods such as elastic weight consolidation \cite{Kirkpatrick2016OvercomingCF}, rehearsal strategies \cite{Rolnick2018ExperienceRF}, and modular architectures \cite{Rusu2016ProgressiveNN}. While much of this work focuses on performance stability, recent efforts have begun to explore efficiency-oriented CL, minimizing retraining costs while maximizing knowledge retention \cite{DeLange2019ACL}.

This paradigm shift is made feasible by recent progress in active learning, online learning, and few-shot or meta-learning methodologies \cite{Vettoruzzo2023AdvancesAC, Santosh2023ActiveLT, Singh2025PATLPA}. These techniques allow models to generalize effectively from limited data and to update incrementally as new information becomes available, mirroring the human approach to learning. Formally, this continuous refinement can be expressed as:
\begin{align}
&M_{t+1} = f(M_t, D_t),
\end{align}
where \(M_t\) denotes the model state at time \(t\), \(D_t\) represents newly observed data, and \(f(\cdot)\) is an update function that integrates the new information into the existing model. Rather than retraining on large static datasets, the model evolves over time, incorporating new knowledge as it emerges.

This daily learning framework improves responsiveness, accuracy, system transparency, and trustworthiness. Incremental learning allows for continuous auditing, timely correction, and gradual refinement, which are characteristics essential for building ethical, explainable, and accountable AI systems. Technically, this can be viewed as a form of `agreeing to disagree' \cite{Bouguelia2017AgreeingTD}, where the model defers to human judgment when uncertainty is high, which facilitates collaborative error correction. Ultimately, learning every day promotes data and energy efficiency, and a more humane and sustainable trajectory for AI development.

This paper builds upon such work to argue for HAI, hybrid architectures where AI models learn incrementally, guided by human feedback and constrained by real-world energy costs. Drawing inspiration from biological cognition, such systems embody both adaptive generalization and resource frugality, key features for an environmentally viable AI ecosystem.

\subsection{Meta-Learning and Data Efficiency}
Meta-learning, or `learning to learn,' seeks to develop models that can adapt to new tasks using minimal data. Pioneering works such as MAML \cite{Finn2017ModelAgnosticMF} and Reptile \cite{Nichol2018OnFM} demonstrate that models can acquire inductive biases across tasks, leading to strong few-shot generalization. More recent studies \cite{Rusu2018MetaLearningWL, Hospedales2020MetaLearningIN} explore gradient-based and metric-based meta-learning for domains ranging from robotics to NLP.

The relevance of meta-learning to sustainable AI lies in its capacity to minimize data and computational overhead. Instead of re-training models from scratch, meta-learners quickly adapt using task-specific information, thereby reducing both time and energy expenditure. When combined with selective data acquisition techniques, meta-learning can drastically reduce the total training footprint without compromising performance.

Importantly, some recent work \cite{Antoniou2018HowTT} shows that task diversity, not data volume, is the key determinant of generalization in meta-learners, providing empirical support for the notion that smaller, high-quality datasets may be more beneficial than large-scale, redundant corpora.

\subsection{Human-in-the-Loop as Governance Mechanism}\label{sec:HITL}
Traditional AI systems are trained on static, often unvetted data, and operate without human oversight post-deployment. This introduces serious risks in high-stakes domains (e.g., healthcare, disaster response, law enforcement). HAI offers a systemic inversion: placing humans \textit{inside} the learning loop, not just as annotators, but as \textit{judicious stewards of model adaptation}.

HITL systems integrate human judgment into the learning process, offering a promising route toward more efficient, ethical, and explainable AI systems. As early as \cite{Amershi2014PowerTT}, interactive ML frameworks showed that non-expert users could significantly enhance model performance by correcting misclassifications or guiding data collection.

HITL methods are particularly effective in data-sparse, high-stakes domains such as healthcare \cite{Holzinger2016InteractiveML}, bioinformatics, and epidemic modeling \cite{Reich2022CollaborativeHM}. By focusing computational resources on informative or uncertain examples, often identified through active learning techniques \cite{Thomas2024ImprovingAL}, these systems reduce the need for exhaustive labeling and model retraining.

Recent work also explores budget-aware and carbon-aware variants of active learning \cite{Konyushkova2017LearningAL, Thomas2024ImprovingAL}, where sample selection is constrained by energy usage or inference latency. These advances directly support the case for integrating HITL mechanisms into a sustainable AI pipeline, particularly in time-sensitive, high-uncertainty scenarios such as pandemics or disaster response.

HITL interaction is not intended as continuous micromanagement, but rather as a strategic intervention triggered when the model's uncertainty or the potential cost of error surpasses a learned threshold. This approach supports rapid decision-making under data scarcity such as during emerging epidemics, while enabling governance-by-design, where human oversight actively shapes model adaptation. Additionally, it facilitates the generation of auditable decision traces, ensuring compliance with legal and ethical frameworks such as the GDPR and the EU AI Act.

Future AI governance frameworks should treat human-AI interaction as a formalized control layer, not an afterthought or interface feature.

\subsection{Toward Human-Centered Trustworthy AI: Unifying Active Learning and Explainability}

Sustainability in AI is not only a matter of energy efficiency or carbon metrics; it also encompasses epistemic integrity and human alignment. In this context, \emph{trustworthy AI} becomes a necessary pillar of sustainable intelligence: models must be not only efficient but also intelligible, correctable, and accountable. We argue that trustworthiness emerges most robustly from the synergy between two often-separate research domains: AL and explainable AI (XAI). Together, they enable systems that learn responsibly, adaptively, and in alignment with human values.

\subsubsection{Human-in-the-Loop Trust through Active Learning}
AL provides a natural pathway toward trustworthy AI by maintaining humans in the training loop (see Section ~\ref{sec:HITL}). Instead of training models passively on fixed datasets, AL dynamically selects informative samples based on model uncertainty or disagreement~\cite{settles2009active,Santosh2023ActiveLT}. This ensures that the model continuously learns from edge cases where human expertise is most valuable. Beyond efficiency, this process establishes a feedback mechanism that enforces accountability: every queried instance can be explained, audited, and justified.

Mathematically, if $f_\theta(x)$ denotes the model parameterized by $\theta$, active learning optimizes information gain by querying samples $x^*$ that maximize the expected reduction in model uncertainty:
\[
x^* = \arg\max_{x \in \mathcal{U}} \mathbb{E}_{y \sim f_\theta(x)}[H(y) - H(y | x)],
\]
where $H(\cdot)$ denotes entropy. Coupling this querying process with human validation embeds an interpretability checkpoint, ensuring that model evolution is transparent and aligned with domain reasoning.

\subsubsection{Explainability Beyond Visual Plausibility}
XAI has conventionally prioritized visual interpretability through saliency maps, attention heatmaps, and feature attributions~\cite{RibeiroAssociationFC, Lundberg2017AUA}. These methods are crucial for communication but limited for trust: they explain \emph{what} the model sees, not \emph{how} it learns. Post-hoc XAI techniques often produce visually plausible results that may not correspond to actual decision mechanisms~\cite{Adebayo2018SanityCF}. To build genuine trust, explanations must be embedded within the model itself.

Integrating explainability into training yields two key benefits. First, it constrains learning toward semantically stable features, improving adversarial and out-of-distribution robustness~\cite{Wang2025ExplainabilityDrivenDG}. Second, it enables continuous interpretability, where every model update through active learning is explainable by design. Such integration transforms explainability from a retrospective diagnostic into a proactive design principle.

%\subsubsection{Synergizing AL and XAI for Trustworthy AI}
%The synergy between AL and XAI creates a self-reinforcing cycle of learning and validation. Explanations guide active sampling by identifying uncertain or ambiguous regions in feature space, while AL ensures that new data continuously refine the explanatory scope of the model. For instance, when an attribution map reveals spurious correlations, AL can prioritize acquiring samples that challenge those biases, closing the loop between interpretability and reliability.

%This continuous explanation-driven learning cycle aligns with the HAI vision of incremental, adaptive, and responsible learning. Each iteration not only improves model accuracy but also strengthens transparency and reduces environmental and cognitive waste by minimizing retraining overhead. The resulting AI system is therefore not just energy-efficient, but also \emph{epistemically efficient}, capable of explaining and justifying its evolution over time.

%\subsubsection{Toward Continuous and Accountable Intelligence}
A truly trustworthy AI is one that learns continuously, explains coherently, and acts responsibly. By embedding explainability into AL loops, models become inherently auditable, enabling trust through traceable updates and interpretable behavior. Future sustainable AI frameworks must therefore integrate active learning, XAI, and ethical governance into a unified paradigm: one where explanations drive learning, learning refines explanations, and both evolve toward shared human and societal goals.

\subsection{The Cognitive Shift: From Static to Lifelong AI}
Conventional deep learning systems treat learning as a single-shot optimization, after which the model is frozen. In contrast, human cognition is incremental, context-sensitive, and resource-aware. HAI operationalizes these cognitive principles by supporting a task-aware memory consolidation, adaptation under strict carbon and annotation budgets, and graceful forgetting mitigation without full retraining. This marks a step toward embodied, cognitively inspired AI, where the system adapts like a human: learning small, important things each day, rather than periodically ingesting terabytes of redundant data.

HAI philosophical implication is a shift in how we define intelligence. It reframes `intelligence' from the compression of vast data to the efficient and context-sensitive adaptation under constraints under real-world constraints. This reframing aligns more closely with human cognition, where intelligence is often demonstrated through timely, resource-aware decision-making rather than brute-force data processing.

% This process embodies the core of ``Human AI''---machines that learn and adapt like humans. By processing small amounts of information regularly, AI systems avoid the inefficiency and environmental cost of large-scale retraining, while staying responsive to dynamic real-world conditions.

% Furthermore, this incremental learning approach improves transparency and trust. Models that evolve gradually can be audited, corrected, and aligned with human values continuously, making AI systems more ethical, accountable, and sustainable.

\section{Where Computational Power Should Matter}
AI does not need to be computationally intensive in all scenarios. In time-critical systems, such as autonomous vehicles (e.g., Tesla), microsecond-level latency is essential. In these high-stakes environments, the computational investment required for real-time inference and rapid decision-making is both necessary and justified. However, the majority of AI applications do not operate under such extreme temporal constraints. In domains like public health, education, environmental monitoring, and financial services, learning in small, daily increments is not only sufficient but often preferable. Incremental learning strategies in these contexts can enhance adaptability, lower infrastructure costs, and significantly reduce energy consumption and associated carbon emissions.

This contrast suggests a fundamental principle: computational power should be allocated strategically, not by default. By concentrating intensive resources where responsiveness is mission-critical, and adopting lightweight, sustainable learning mechanisms elsewhere, we can construct an AI ecosystem that balances performance with environmental and societal responsibility. Efficiency, in this framing, is not merely a constraint; it is a design objective.

\section{The Path Forward: Responsible Human-Like AI}
The premise of AI has traditionally been to mimic human intelligence; however, it should instead aim to augment human intelligence. Contemporary AI models predominantly replicate human outputs without capturing the underlying cognitive processes that generate them. Human cognition is distinguished by the ability to generalize from limited experience, reason effectively under uncertainty, and continuously refine understanding through ongoing interaction with the environment. 

In contrast, as discussed previously, most current AI systems rely on static, large-scale datasets and are retrained in isolated cycles, often detached from the contextual and temporal dynamics of real-world environments. This fundamental divergence highlights the urgent need to realign AI development with the principles of human learning to enhance adaptability, relevance, and ethical integration.

Achieving this realignment requires designing AI models that learn in small, meaningful increments, daily, contextually, and responsibly. Moreover, active learning, explainable AI, and robust ethical frameworks \cite{Santosh2024CrackingTM,santosh2022ai,Wall2025WinsorCAMHV,Wang2025ExplainabilityDrivenDG} are no longer optional components of AI development. Without continuous learning, AI models cannot effectively adapt to evolving real-world dynamics. Without transparency and ethics, AI risks perpetuating harm rather than promoting benefit. 

HAI represents the crucial bridge between intelligence and responsibility, combining adaptability with accountability to forge the next generation of sustainable, human-centric AI systems.

%To realign AI with human cognition, we must build systems that learn in small, meaningful steps - daily, contextually, and ethically. This brings us to the foundation of this vision.

% \begin{quote}
% \textbf{Human-centered AI, often referred to as Human AI, emphasizes the design and development of artificial intelligence systems that prioritize human values, ethics, and societal well-being. Rather than focusing solely on performance or automation, this approach ensures that AI technologies are transparent, explainable, and aligned with human intentions. It advocates for inclusive participation in AI development, considering diverse perspectives and minimizing bias. By fostering trust, accountability, and empathy in machine behavior, Human-centered AI aims to create systems that not only serve but also respect and empower individuals and communities.}
% \end{quote}

%This definition serves as a compass for how we should design and deploy AI systems in the future.

\begin{figure}[tbp]
    \centering
    \includegraphics[width=0.7\linewidth]{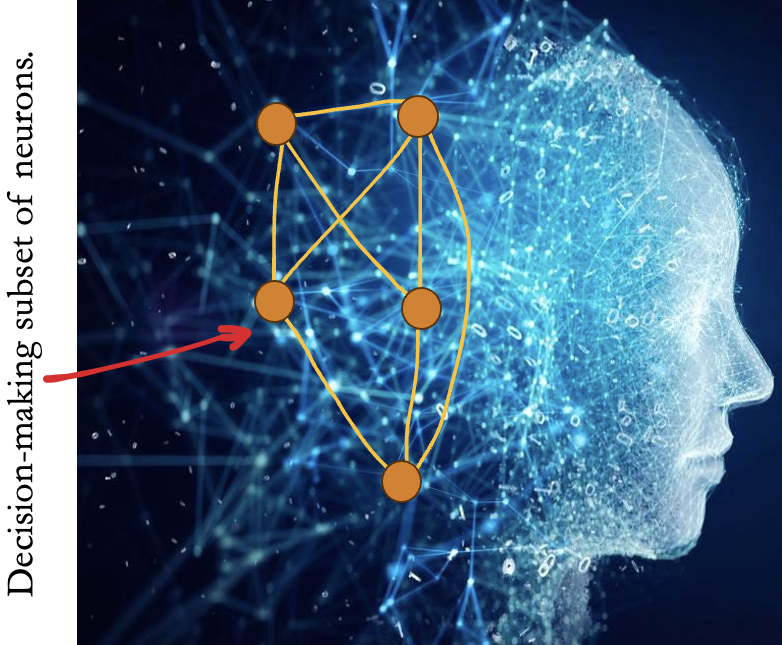}
    \caption{Humans selectively engage a relevant subset of neurons based on the nature and complexity of the input.}
    \label{fig:humanAI}
\end{figure}
\section{Toward Adaptive and Energy-Efficient Artificial Neural Network Architectures}

\begin{figure}[tbp]
    \centering
    \includegraphics[width=0.7\linewidth]{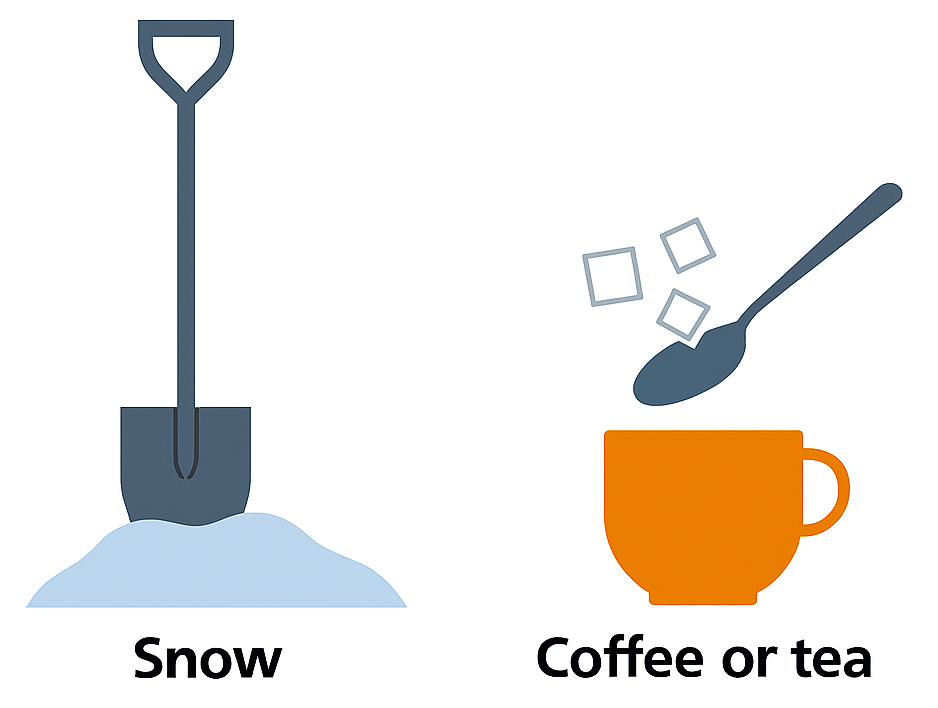}
    \caption{Illustration of task-specific resource allocation using a tool analogy. Just as a shovel is suited for moving snow and a spoon for adding sugar to coffee or tea, neural systems should engage distinct subsets of neurons depending on the nature and complexity of the input. This analogy underscores the importance of selective activation in efficient cognitive processing.}
    \label{fig:analogy}
\end{figure}

Although the human brain contains billions of neurons, it does not activate all of them for every task. Instead, it selectively engages a relevant subset based on the nature and complexity of the input \cite{wang2025pcsnnpredictivecodingbasedlocal}. Human intelligence operates by recruiting only the necessary neural resources, enabling efficient cognitive processing without imposing excessive computational burden (see Figure~\ref{fig:humanAI}). This selective activation is especially effective for routine or low-complexity tasks.  An apt analogy (Figure~\ref{fig:analogy}) is knowing when to use a shovel versus a spoon: a shovel is perfect for moving snow, while a spoon is ideal for adding sugar to coffee or tea (though, of course, sugar ruins the taste of both!). Just as each tool suits a different purpose, neural resources should be engaged selectively depending on task demands. There is no need to reach for a deep learning model (the proverbial shovel) when a lightweight or shallow model (the spoon) can get the job done. Matching model complexity to task complexity is not only computationally efficient but also a step toward more sustainable and interpretable AI systems.

In contrast, most artificial neural networks remain static in both structure and behavior, activating millions of parameters uniformly regardless of task demands. For simple queries, the use of hundreds of layers and millions of neurons is computationally excessive and environmentally unsustainable. This inefficiency underscores a fundamental mismatch between biological and artificial intelligence.

To address this gap, as mentioned before, AI systems must adopt dynamic, task-sensitive activation strategies that mirror the selective efficiency of human cognition. One promising direction is integrating liquidity with neural networks \cite{Hasani2020LiquidTN}, which adapt their internal dynamics in response to input complexity. Unlike traditional architectures, liquid networks adjust their computational pathways in real time, allowing for more efficient processing and better generalization across tasks. This flexibility reduces energy consumption while enhancing interpretability and responsiveness, which are key attributes for sustainable and ethical AI.

By activating only the neurons required for a given task, these types of neural networks offer a biologically inspired alternative to static deep learning models. They represent a critical step toward building AI systems that are both cognitively aligned and environmentally responsible.

\section{Problem Formulation}\label{sec:formulation}
This section establishes formal definitions for the objectives of the proposed Human AI paradigm, introducing the relevant notation, constraints (e.g., carbon budgets), and optimization goals. The goal is to reframe sustainable AI not merely as an engineering concern but as a constrained learning optimization problem, suitable for theoretical and empirical investigation.

\subsection{Task-Based Formulation}
Let $\mathcal{T} = \{ \mathcal{T}_1, \mathcal{T}_2, \ldots, \mathcal{T}_n \}$ denote a distribution over tasks, where each task $\mathcal{T}_i$ is associated with a data distribution $\mathcal{D}_i$, and a learning objective $\mathcal{L}_i(\theta)$, parameterized by model weights $\theta$. In classical machine learning, the goal is to learn parameters $\theta^*$ that minimize the expected loss over this task distribution:
\begin{align}
& \theta^* = \arg\min_{\theta} \mathbb{E}_{\mathcal{T}_i \sim \mathcal{T}} \left[ \mathcal{L}_i(\theta) \right].
\end{align}
However, in the context of sustainable, human-centered AI, this objective is subject to several real-world constraints that are rarely modeled explicitly:

\subsection{Carbon-Aware Learning Constraints}\label{subsec:carbon_constraints}
Let $C(\theta, \mathcal{T}_i)$ represent the estimated carbon cost (e.g., in kg CO$_2$e) incurred when training or adapting model $\theta$ on task $\mathcal{T}_i$. Our reformulated optimization must satisfy: %\todo{need to check whether it is complete.}
\begin{align}
\text{subject to } C(\theta, \mathcal{T}_i) \leq \epsilon \quad \forall \mathcal{T}_i,
\end{align}
where $\epsilon$ is a carbon budget, a task-specific or global constraint reflecting ecological boundaries (e.g., carbon-neutral policies, data center limits). Estimating $C$ may involve proxies such as FLOPs, runtime, memory usage, or hardware type.

\subsection{Data-Efficiency and Human Interaction Budget}
Let $D_i \subset \mathcal{D}_i$ denote a labeled subset of task data curated through active or HITL selection. The \textit{information gain per labeled example} is critical. We introduce a function $U(x; \theta)$ measuring the uncertainty or expected informativeness of an unlabeled instance $x$, and formulate a budgeted active learning loop:
\begin{align}
D_i = \arg\max_{|D_i| \leq b} \sum_{x \in D_i} U(x; \theta),
\end{align}
where $b$ is a human annotation budget or attention span constraint. This reflects the principle that \textit{not all data is equally valuable}, especially under ecological and temporal constraints.

\subsection{Continual Adaptation without Forgetting}
Let $\theta_t$ denote model parameters after learning on task $\mathcal{T}_t$. We require that performance on prior tasks $\mathcal{T}_k$, $k < t$, remains within a tolerable degradation margin $\delta$:
\begin{align}
\mathcal{L}_k(\theta_t) - \mathcal{L}_k(\theta_k) \leq \delta \quad \forall k < t.
\end{align}
This constraint models catastrophic forgetting mitigation and aligns with continual learning paradigms. Unlike traditional learning where models are retrained globally, our objective promotes \textit{local plasticity under global stability}, minimizing retraining overhead.

\subsection{Overall Objective}
We now define a multi-objective constrained optimization problem:
\begin{align}\nonumber
&\min_{\theta} \quad \mathbb{E}_{\mathcal{T}_i \sim \mathcal{T}} \left[ \mathcal{L}_i(\theta) \right] + \lambda \cdot R(\theta)\\
&\text{subject to } C(\theta, \mathcal{T}_i) \leq \epsilon, \quad |D_i| \leq b, \quad \Delta \mathcal{L}_k \leq \delta,
\end{align}
where $R(\theta)$ is a regularization term reflecting model size or energy footprint, $\lambda$ balances predictive performance with sustainability, and constraints encode ecological, human, and cognitive limits.

\subsection{From Optimization to Architecture}
This formulation serves as a design principle for constructing \textit{HAI systems} that balance:
\begin{itemize}
    \item \textit{Sustainability} (via carbon constraints and efficiency metrics),
    \item \textit{Adaptability} (via few-shot/meta-learning),
    \item \textit{Robustness} (via continual learning), and
    \item \textit{Human alignment} (via selective supervision and explainability).
\end{itemize}
In what follows, we instantiate this framework in a modular system that fuses meta-learning cores, energy-aware controllers, and interactive human interfaces.

\section{Human AI: A Proof-of-Concept}
\subsection{Conceptual Architecture}
We propose HAI, a hybrid carbon-aware learning architecture that integrates meta-learning cores, active data selection, continual adaptation, and HITL feedback under strict energy and annotation budgets. HAI departs from monolithic, one-shot models by operating in a lifelong, incremental, and budget-constrained scenario.

\begin{figure*}[tbp]
    \centering
    \includegraphics[width=0.8\textwidth]{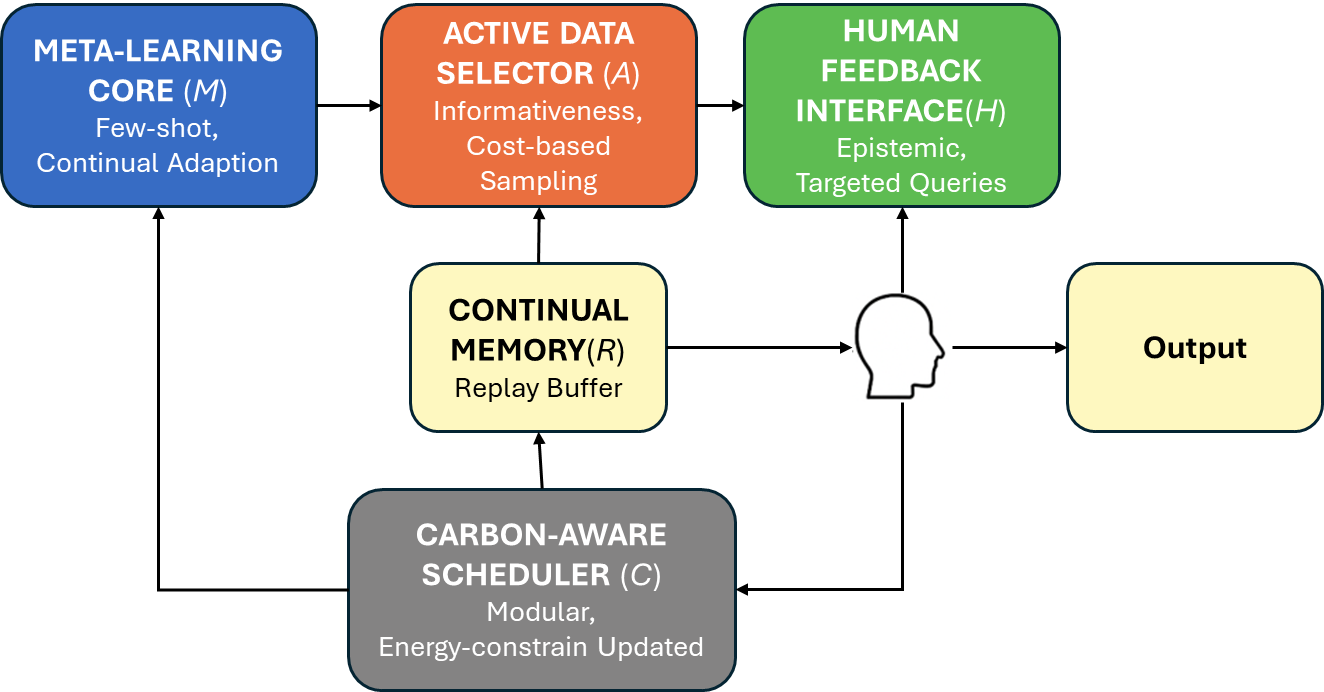}
    \caption{Human AI (HAI) modular architecture, comprising Meta-Learning Core ($\mathcal{M}$), Active Data Selector ($\mathcal{A}$), Carbon-Aware Scheduler ($\mathcal{C}$), Human Feedback Interface $(\mathcal{H})$, and Continual Memory $(\mathcal{R})$. Arrows denote data and control flow across modules under energy and annotation constraints.}
    \label{fig:architecture}
\end{figure*}

The HAI architecture consists of the following core modules (see Fig.~\ref{fig:architecture} for an overview):
\begin{itemize}
    \item \textit{Meta-Learning Core} ($\mathcal{M}$): A parameter-efficient backbone trained to rapidly adapt to new tasks using few-shot supervision. This module leverages prior experience over a distribution of tasks to initialize weights with strong inductive biases.
    
    \item \textit{Active Data Selector} ($\mathcal{A}$): A learned acquisition function that scores unlabeled samples using informativeness (e.g., entropy, BALD) and cost (e.g., annotation time, energy). It selects the most valuable data points under budget $b$.
    
    \item \textit{Carbon-Aware Scheduler} ($\mathcal{C}$): Tracks and optimizes energy consumption by dynamically selecting compute paths (e.g., shallow adapters vs full finetuning), prioritizing low-FLOP updates and offloading to green energy windows when available.
    
    \item \textit{Human Feedback Interface} ($\mathcal{H}$): Provides a visual explanation interface and receives targeted human input (e.g., label, correction, ranking). It supports epistemic uncertainty estimation, helping humans guide model correction rather than labeling exhaustively.
    
    \item \textit{Continual Memory} ($\mathcal{R}$): A memory buffer with selective rehearsal, storing exemplars and adaptation metadata. It mitigates forgetting and facilitates periodic replay under compute and storage budgets.
\end{itemize}

\subsection{Data Efficiency through Human-AI Collaboration}
At the heart of HAI lies the principle that \textit{human knowledge is costly but critical}. Rather than relying on passively collected large datasets, HAI actively queries the human only when the expected information gain per query justifies the cost. We define the utility of acquiring label $y$ for input $x$ as an informativeness function $U(x; \theta)$:
\begin{align}
U(x; \theta) = H[p(y|x; \theta)] + \beta \cdot \text{Var}_{\theta \sim q(\theta)}[p(y|x; \theta)],
\end{align}
where  $H[\cdot]$ is Shannon entropy (epistemic uncertainty), $q(\theta)$ is posterior over model weights, and $\beta$ is trade-off parameter for model confidence vs disagreement.

Samples with highest $U(x; \theta)$ are selected under the budget $b$. Unlike classical active learning, HAI uses a \textit{dynamic query strategy}, adjusting sampling frequency based on human availability, context urgency (e.g., in pandemics), and compute energy states.

HITL operations in HAI are designed to maximize the value of limited human attention. These interactions include direct labeling for samples where the model exhibits high uncertainty, correction or confirmation for low-confidence predictions, and rule injection, where humans contribute constraints or logical rules to guide inference, such as through programmatic supervision or weak labeling. This collaborative mechanism enables the system to incorporate human expertise efficiently without requiring exhaustive annotation.

\subsection{Carbon-Aware Learning Mechanisms}

To enforce carbon constraints (see Section~ \ref{subsec:carbon_constraints}), HAI employs \textit{adaptive training and inference strategies} through an energy profiling layer.

Each learning operation is tagged with an estimated FLOP and wattage cost, derived from profiling tools (e.g., CodeCarbon, Nvidia Nsight). A cumulative energy budget tracker estimates:\(
C_t = \sum_{i=1}^t E(\theta_i, \mathcal{T}_i),
\)
where $C_t \leq \epsilon$ for all $t$. During high-load or carbon-intensive times (e.g., peak energy hours), HAI postpones compute-heavy updates, prioritizes cache lookups, and invokes \textit{shallow model pathways}.

\subsection{Toward Responsible, Systems-Level AI}
HAI is more than a model; it is a learning system with built-in accountability and sustainability constraints. It can serve as a template for regulatory-compliant AI tools (e.g., in healthcare or finance), green-by-default ML toolkits (via adapterized architectures), and ethical deployment frameworks for governments and NGOs.

We advocate a broader shift in ML research from model-centric to systems-centric design, where resources, stakeholders, and governance mechanisms are integrated into the design loop, not retrofitted afterward.

\section{Final Reflection: Unified benchmark for Sustainable AI}
We aim at creating a standardized benchmark suite that jointly evaluates models on accuracy, energy usage, carbon impact, and human annotation cost, encouraging the research community to embrace multi-objective optimization by default.  We propose that future ML benchmarks include energy-aware performance metrics and require carbon reporting alongside accuracy. We suggest the adoption of the \textit{carbon-accuracy tradeoff curve} as a quantitative tool to guide model selection in real-world deployment, particularly where energy or emissions are constrained (e.g., mobile devices, developing regions, or climate-aware enterprises).

\section{Conclusion and Takeaways}
This paper highlights the urgency of rethinking AI development through the lens of sustainability, adaptability, and human alignment. The conventional emphasis on big data and monolithic training cycles not only escalates computational costs and carbon emissions but also limits AI’s responsiveness to dynamic, real-world conditions. By adopting principles inspired by human cognition, i.e., continuous, incremental learning under explicit carbon and annotation budgets, we propose a paradigm shift embodied in the Human AI (HAI) framework. HAI integrates meta-learning, active human collaboration, and energy-aware adaptation to create AI systems that are both effective and environmentally responsible.  It is a modular, carbon-aware, and human-aligned learning framework that reconceptualizes AI development for the age of environmental and ethical urgency. We formulated the core problem as a multi-objective constrained optimization, balancing predictive performance with energy budgets, memory retention, and limited human labor.
\begin{quote}
{\em ``The environmental cost of AI is not an unfortunate side-effect; it is a solvable design flaw.''}
\end{quote} 
Through HAI, we have shown that it is possible to build learning systems that are efficient, responsible, and fundamentally human-centered. As AI becomes more embedded in societal infrastructure, its alignment with ecological boundaries and democratic control must be engineered from the ground up and not added later.
The future of AI is not just about smarter models; it is about wiser systems.

%Further, by exploring biologically inspired architectures, we pave the way for models that dynamically allocate computational resources, matching task complexity with energy efficiency. Ultimately, bridging the gap between intelligence and responsibility requires embedding human values and ecological constraints into AI design, advocating for technologies that serve society sustainably and ethically.

%\textbf{Key Takeaways:}
% \begin{itemize}
%     \item Sustainability is tractable. Performance does not require energy-intensive overfitting. Structured modular updates paired with energy-aware scheduling are highly effective.
%     \item Human feedback is not an overhead; it is a governance layer. When used strategically, humans provide targeted corrections that boost learning and align models with societal norms.
%     \item Intelligence under constraints is more realistic and human-like than unconstrained scaling. We must evolve beyond data and compute maximalism toward cognitive minimalism.
% \end{itemize} 

%\textbf{Key Takeaways:} 
In short, the following are the key takeaways. Sustainability in AI is not only achievable, but tractable: (a) performance does not necessitate energy-intensive overfitting. Instead, structured modular updates, when combined with energy-aware scheduling, offer a highly effective alternative. (b) Human feedback should not be viewed merely as overhead; rather, it serves as a vital governance layer. When applied strategically, human interventions offer targeted corrections that improve learning efficiency and help align AI models with broader societal norms. Finally, (c) intelligence developed under constraints may, in fact, be more realistic and human-like than models derived from unconstrained scaling. To move forward, the field must shift away from data and compute maximalism and toward a philosophy of cognitive minimalism that emphasizes efficiency, adaptability, and purpose-driven learning.

%The future of AI must not be driven solely by data scale or computational horsepower. It must be guided by sustainability, ethics, and a deep respect for how humans actually learn and grow. We do not solve real-world problems by waiting for enough data. We solve them by learning continuously, like humans do. This is how we reduce carbon footprints, increase responsiveness, and make AI an enduring ally in facing inevitable global challenges -- from pandemics to climate change. {\em Do not wait for data. Learn every day.} That is not just a call to action -- it is the foundation for a new era of artificial intelligence: adaptive, ethical, explainable, and human.
\balance
\bibliographystyle{IEEEtran}
\bibliography{references}

\end{document}